\documentclass[letterpaper, 10 pt, conference]{ieeeconf} % Comment this line out if you need a4paper

\usepackage{mathtools}
\usepackage[listings,skins]{tcolorbox}
\usepackage{booktabs}
\usepackage{amssymb}

\usepackage{float}
\usepackage{makecell}
\usepackage{calc}
\usepackage{caption}
\usepackage{subcaption}

\usepackage[font=small,labelfont=bf]{caption}
\usepackage{subcaption}
\usepackage{booktabs}
\usepackage{makecell}
\usepackage{graphicx}
\usepackage{placeins}

\usepackage{colortbl}

\usepackage[
  style=ieee,
  backend=biber,
  sorting=none,
  sortcites=true
]{biblatex}

\usepackage{multirow}
\usepackage{multicol}
\usepackage{amsthm}
\usepackage{url}

\usepackage{colortbl}

\definecolor{gr}{rgb}{0.921, 0.972, 0.905}
\definecolor{pink}{rgb}{0.972, 0.905, 0.917}
\definecolor{redishh}{rgb}{0.9, 0.17, 0.31}
\definecolor{redish}{rgb}{1.0, 0.01, 0.24}
\definecolor{antique}{rgb}{0.57, 0.36, 0.51}
\definecolor{darkcandy}{rgb}{0.64, 0.0, 0.0}
\definecolor{pastel}{rgb}{0.09, 0.45, 0.27}

\IEEEoverridecommandlockouts        % This command is only needed if 
\title{
MedGate-Fusion: Integrating First-Encounter Semantic Narratives and Physiological Biomarkers for Prospective Stroke Risk Stratification

}

\author{Hemn Khdr$^1$, Mohammad Noaeen$^1$, Karim Keshavjee$^1$, Aziz Guergachi$^2$, Zahra Shakeri$^{1,3,4}$\\
\textit{$^1$Dalla Lana School of Public Health, University of Toronto, Toronto, Canada}\\
\textit{$^2$Department of Information Technology Management, Toronto Metropolitan University, Toronto, Canada}\\
\textit{$^3$ Faculty of Information, University of Toronto, Toronto, Canada}\\
\textit{$^4$ Schwartz Reisman Institute, University of Toronto, Toronto,  Canada}
}

\begin{document}
\maketitle
\thispagestyle{empty}
\pagestyle{empty}

%%%%%%%%%%%%%%%%%%%%%%%%%%%%%%%%%%%%%%%%%%%%%%%%%%%%%%%%%%%%%%%%%%%%
\begin{abstract}
Prospective stroke risk stratification in primary care is challenging because early risk signals are distributed across routine biomarkers and unstructured clinical narratives. We propose \texttt{MedGate-Fusion}, a multi-modal gated architecture that integrates transformer-based embeddings of first-encounter narratives with ten routinely recorded risk markers. We used electronic medical record data from the Canadian Primary Care Sentinel Surveillance Network (CPCSSN). Starting from 808,921 encounter-level observations, we constructed a first-encounter cohort and retained 102,736 unique patient records with non-empty narratives and sufficient data to evaluate a five-year stroke outcome. To reduce explicit target leakage from diagnostic mentions in notes, we applied dictionary-based redaction of stroke-related terms prior to semantic encoding.
\end{abstract}

%----------------------------------
\definecolor{amaranth}{rgb}{0.9, 0.17, 0.31}
\definecolor{gr}{rgb}{0.55, 0.71, 0.0}
\definecolor{ashgrey}{rgb}{0.7, 0.75, 0.71}
%----------------------------------

%%%%%%%%%%%%%%%%%%%%%%%%%%%%%%%%%%%%%%%%%%%%%%%%%%%%%%%%%%%%%%%%%%%%%%%%%%%%%%%%

\section{Introduction}

Stroke is a leading cause of long-term disability and mortality, and its risk is elevated in individuals with cardiometabolic conditions such as hypertension and type 2 diabetes mellitus \cite{tsao2023heart,feigin2021global}. Early identification of high-risk patients in primary care is therefore a high-impact opportunity for prevention \cite{Sarwar2010}. However, prospective risk estimation is difficult because clinically relevant signals are distributed throughout both structured measurements (e.g., blood pressure, lipids, glycemic markers) and unstructured narratives recorded during routine visits \cite{sheikhalishahi2019natural}.

To enable timely and life-saving clinical interventions in primary care settings, clinicians require accurate risk identification tools. Current risk calculators often ignore the rich and longitudinal narratives that exist in electronic medical records \cite{DAgostino2008,liu2025estimating,jonnagaddala2015coronary}. This creates a need for advanced systems that can process the massive data generated during routine patient visits. Therefore, the development of predictive models is essential to reduce the global burden of diabetic complications. Electronic medical records store a high and unique density of physiological biomarkers and unstructured clinical narratives \cite{jensen2012mining,nerella2024transformers}. These digital repositories contain the hidden and early signals of vascular deterioration that standard screening methods miss. To identify subtle patterns across thousands of patient records, predictive algorithms process diverse and dense data sources.

Prior studies show that machine learning can improve cardiovascular risk prediction from routine structured data \cite{subramani2023cardiovascular,Poplin2018,Weng2017}.
Deep learning can also learn nonlinear patterns across diverse EHR and imaging inputs \cite{nguyen2019predicting,acosta2022multimodal,rajkomar2018scalable}.
However, prospective models still underuse unstructured notes.
Early NLP approaches often reduced documentation to keywords or frequency statistics \cite{sheikhalishahi2019natural}.
Such representations cannot capture negation, timing, or compositional meaning in clinician language.
Clinical and biomedical language models (e.g., ClinicalBERT, BioBERT, PubMedBERT, and large clinical LMs) learn contextual text representations that support richer clinical text mining \cite{alsentzer2019publicly,lee2020biobert,gu2021domain,yang2022large}.
For structured EHR sequences, transformer models such as BEHRT and Med-BERT show that representation learning over longitudinal events improves prediction and can transfer to data-limited settings \cite{li2020behrt,rasmy2021med}.
Even with these advances, prospective stroke prediction that combines dense narratives with physiological measurements remains limited.
Many multimodal fusion studies rely on retrospective designs, inpatient outcomes, or lack controls for label leakage from diagnostic proxies and post-event descriptors in free text \cite{ramadan2025diagnostic,lyu2023multimodal}.

We propose MedGate-Fusion to address this gap by integrating transformer-based semantic embeddings with routinely collected physiological biomarkers under a prospective first-encounter design. This study contributes: (1) a leakage-aware cohort construction strategy, (2) semantic scrubbing of stroke-related terms, and (3) a gated multimodal architecture for five-year stroke risk prediction in primary care.

The remainder of this paper is structured as follows. Section \ref{method} details the methodology of our work and the specific inclusion criteria for the diabetic cohort. This section describes the pipeline for temporal landmark filtering and semantic scrubbing to eliminate target leakage during the evaluation. Section \ref{results} provides the performance results for the proposed fusion architecture and multi-modal ablation studies, followed by Section \ref{discussion} which elaborates on the findings and provides the primary clinical and technical implications of our work. We conclude the paper in Section \ref{conclusion} with a summary of the key findings and future directions of our research.

\section{Methods}
\label{method}

The clinical risk estimation task in this paper is formulated as a prospective and supervised learning objective over a first-encounter clinical manifold.
 For a patient population $\mathcal{P}$, each individual $i$ is represented by a multi-modal feature set $(x_{n,i}, T_i, y_i)$ at their initial documented visit. The numerical vector $x_{n,i} \in \mathbb{R}^{d_n}$ contains the primary physiological biomarkers and physical measurements from the first encounter. The unstructured text $T_i$ provides the diagnostic narrative from the same temporal landmark. The binary variable $y_i \in \{0, 1\}$ denotes the stroke outcome during the subsequent five year follow up period. An optimal mapping function $f: (x_{n,i}, T_i) \to \hat{y}_i$ should minimize a specified loss function $\mathcal{L}$. This prospective design restricts model learning to data recorded before any symptomatic or diagnostic documentation of the event.

\subsection{Data Source and Inclusion Criteria}

\begin{figure*}[!t]
    \centering

    \begin{subfigure}[b]{0.42\textwidth}
        \centering
        \includegraphics[width=\textwidth]{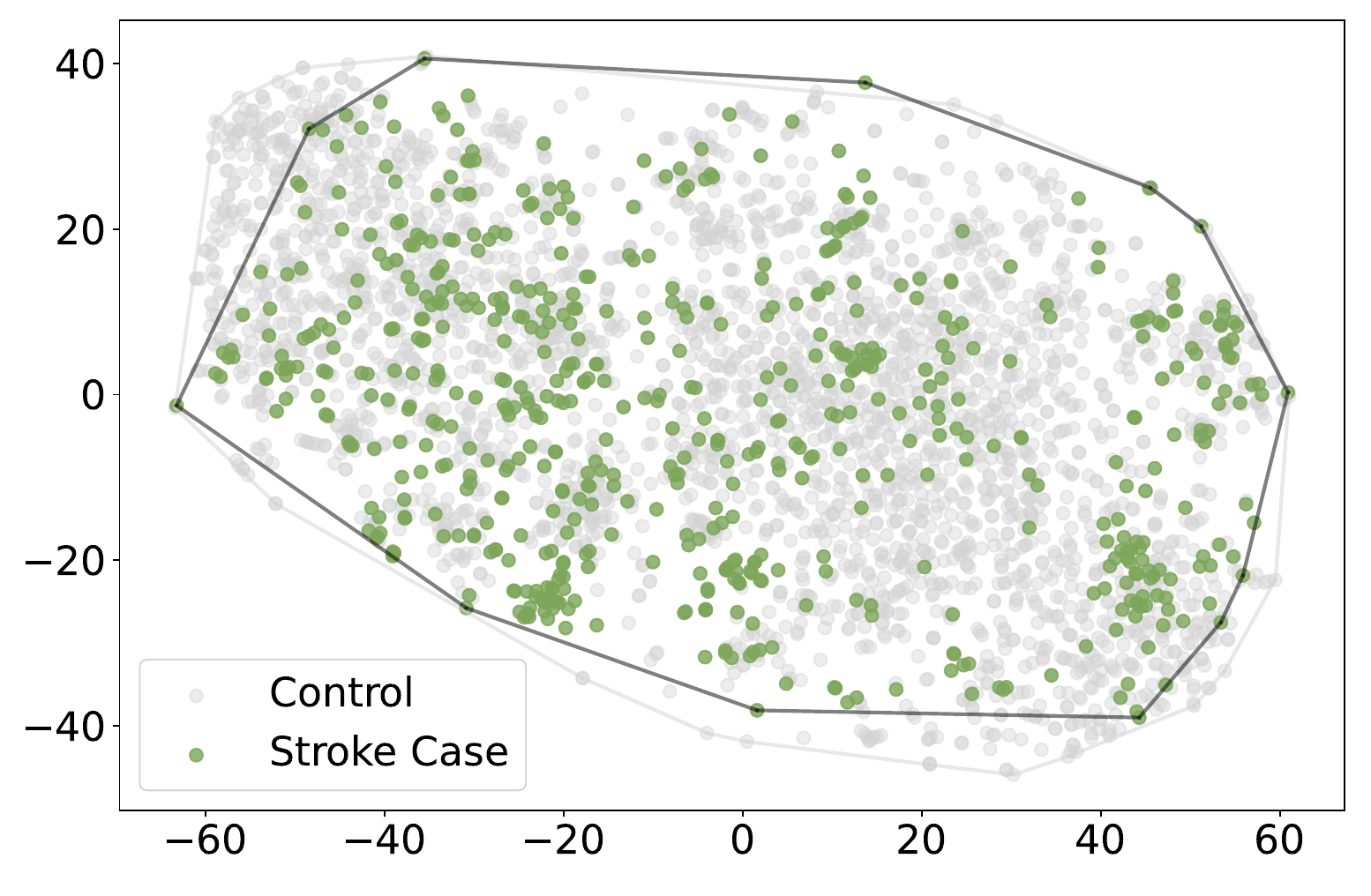}
        \caption{}
        \label{fig:semantic_manifold}
    \end{subfigure}
    \hfill
    \begin{subfigure}[b]{0.45\textwidth}
        \centering
        \includegraphics[width=\textwidth]{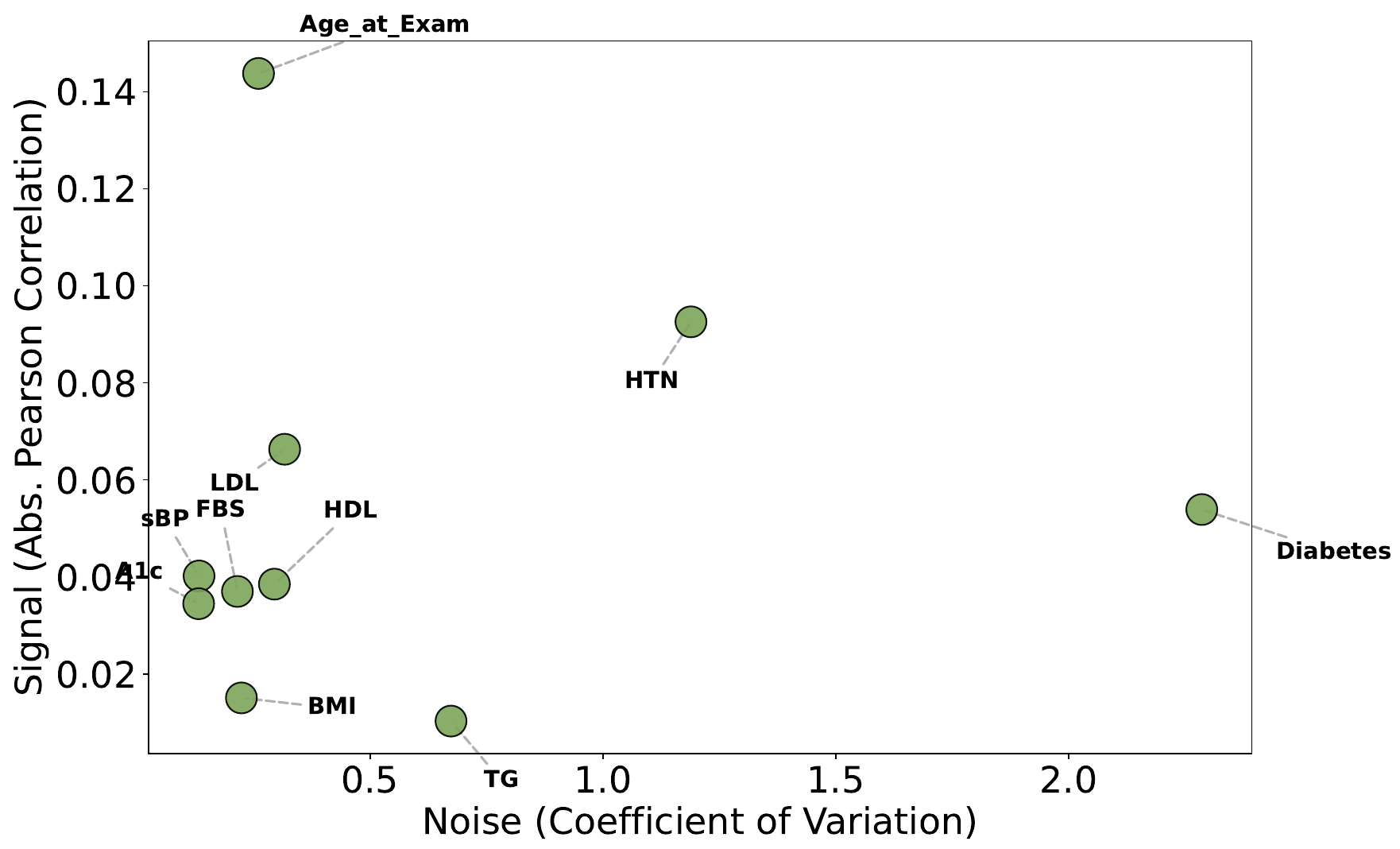}
        \caption{}
        \label{fig:signal_noise}
    \end{subfigure}

    \vspace{-2mm}

    \caption{\small \textbf{Structural Analysis of the First-Encounter Clinical Manifold.} (a) Semantic manifold projection via t-SNE illustrating the high non-linear overlap between stroke and control groups. (b) Annotated signal-to-noise distribution mapping the coefficient of variation against the absolute Pearson correlation for discrete physiological biomarkers.}

    \label{fig:clinical_analysis}

    \vspace{-2mm}
\end{figure*}

We used electronic medical record data from the Canadian Primary Care Sentinel Surveillance Network (CPCSSN) for this prospective cohort study. The initial raw dataset contained 808,921 observations from a diverse and national population of Canadian primary care patients. To establish a true and prospective window, we isolated only the earliest documented data point for each unique and diverse patient. This first-encounter selection limits the model to information available before diagnosis by excluding any later diagnostic entries or downstream clinical documentation. We excluded records without diagnostic narratives in the text fields to preserve parity across modalities. These filtering steps reduced the dataset to a unique set of patient-level records suitable for prospective prediction. The final dataset provides sufficient information density to evaluate the predictive utility of the point-of-care pipeline (Table~\ref{tab:table1}).
 The semantic visualization of t-SNE in Figure~\ref{fig:semantic_manifold} illustrates a high and complex overlap between the stroke and control cohorts. The geometric hulls in the figure specify the boundary regions where semantic narratives provide the primary discriminative signal. This overlap confirms that standard linear modeling is insufficient to resolve early-stage vascular risk factors at the point of contact \cite{jamil2025empirical}. We used deep transformer-based architectures to identify latent and non-linear pathological patterns within the unstructured diagnostic text. Such high-dimensional features capture systemic deterioration signals that traditional biomarkers typically ignore during routine clinical documentation \cite{rajkomar2018scalable,lyu2023multimodal,huang2019clinicalbert, tayefi2021challenges}. Baseline characteristics and cohort structure are summarized in Table~\ref{tab:table1} and Figure~\ref{fig:clinical_analysis}, respectively.

\subsection{Leakage Mitigation and Semantic Scrubbing}

For a valid and predictive evaluation, we designed a dual-layered mitigation strategy to remove direct and indirect target leakage \cite{sasse2025overview}. We first implemented the temporal landmark filter by restricting the input space to the initial visit for each patient. This constraint prevents the model from using rehabilitation narratives or secondary complications as predictive features \cite{singh2015incorporating}. We then identified specific and sensitive keywords such as stroke and cva and infarct that represent the documented outcome. A dictionary-based masking algorithm replaces these occurrences with a neutral $[MASK]$ token during the preprocessing task. The semantic encoder then identifies early signals of vascular deterioration within the initial history. The model therefore emphasizes systemic and indirect patterns rather than the clinical aftermath of the event~\cite{amanatidis2025data}.

\subsection{Transformers for Semantic Representation}

To extract contextual features from clinical narratives, we used the all-MiniLM-L6-v2 sentence-transformer model, which produces compact 384-dimensional embeddings. We selected this model because it provides an efficient balance between semantic representation quality, computational cost, and reproducibility for large-scale 5-fold cross-validation. Larger domain-specific encoders such as ClinicalBERT, BioBERT, and PubMedBERT will be compared in future work. This architecture maps the scrubbed text $T_i$ into a continuous vector space $h_{s,i} \in \mathbb{R}^{384}$. We selected this transformer because it uses self-attention \cite{vaswani2017attention,wang2020minilm} to calculate the contextual relationship between disparate clinical terms. Traditional models treat medical terms as independent tokens and lose the temporal and pathological context. The attention mechanism calculates a weighted sum of values based on the compatibility of a query and key pair:

\begin{equation}
\text{Attention}(Q, K, V) = \text{softmax}\left(\frac{QK^T}{\sqrt{d_k}}\right)V
\end{equation}

The dot product $QK^T$ measures semantic similarity between terms, while $\sqrt{d_k}$ rescales the scores to stabilize gradients during training. This representation captures long and complex dependencies between comorbidities that frequency-based models often ignore. The transformer models the clinical history as a cohesive semantic sequence rather than a bag of words.

\subsection{The MedGate-Fusion Architecture}

We designed the MedGate-Fusion architecture to integrate numerical biomarkers and semantic embeddings through a contextual gating mechanism \cite{wang2025moe,han2024fusemoe}. This structure utilizes two specialized experts to process the different modalities of clinical data. The first branch uses a gradient-boosted ensemble to evaluate numerical risk markers $x_{n,i}$. The second branch processes the semantic transformer embeddings $h_{s,i}$ to identify high and relevant clinical patterns.

To determine the relative contribution of each modality, we implemented a context-aware meta-learner as a gate. Standard concatenation methods assume that the importance of text and lab values is static across all patients. Our approach addresses this limitation by calculating a dynamic weight based on the specific health profile of the individual, similar to recent hybrid architectures (e.g., DeepNote-GNN) that integrate textual embeddings with structured EHR features \cite{he2025comparative}. The gating function $G$ utilizes the raw numerical context and the initial predictions of both experts to calculate a dynamic weight. We define the gating coefficient $\alpha_i$ using a sigmoid activation function:
\begin{equation}
\alpha_i = \sigma(W_g \cdot [p_{n,i}, p_{s,i}, x_{n,i}] + b_g)
\end{equation}
Where $p_{n}$ and $p_{s}$ represent the class probabilities from the numerical and semantic branches, respectively. The final fused prediction $y_{pred,i}$ is a weighted combination of these two specialized signals:
\begin{equation}
y_{pred,i} = (1 - \alpha_i) \cdot p_{n,i} + \alpha_i \cdot p_{s,i}
\end{equation}
This gated design allows the model to prioritize the semantic signal when the numerical lab values are inconclusive or missing. The architecture adapts to the unique and specific informational density of each patient record. To investigate our proposed architecture further, we explored the noise patterns within the numerical biomarkers and the prospective target. The signal-to-noise distribution in Figure~\ref{fig:signal_noise} shows that discrete labs such as A1c and LDL provide a weak predictive signal. As illustrated in this Figure, these biomarkers exhibit high population variance and negligible target correlation compared to the dense information in narratives. The gating function addresses this disparity by dynamically addressing the transformer branch to compensate for the noisy numerical data. This architecture resolves the technical challenge of identifying stroke risk when physiological measurements do not provide a reliable signal. As a result, the model maintains high performance by leveraging the stable semantic context of the patient history.

\subsection{Imbalance Optimization and Objective Function}

To address the extreme class imbalance in the stroke events, we used a weighted binary cross-entropy loss function \cite{salmi2024handling}. Standard loss functions often lead to majority class bias, where the model ignores the rare and critical positive cases. To prevent the model from converging to a trivial solution of predicting the majority class, we utilized cost-sensitive learning. We assigned weights inversely proportional to the class frequencies to provide a high and significant penalty for minority errors. The weight $w_c$ for class $c$ is calculated as $w_c = \frac{N}{K\,n_c}$, where $N$ is the number of samples, $n_c$ is the number of samples in class $c$, and $K=2$ is the number of classes. The loss is
\begin{equation}
\mathcal{L} = -\sum_{i=1}^{N}\left[w_1 y_i \log(\hat{y}_i) + w_0(1-y_i)\log(1-\hat{y}_i)\right].
\end{equation}

This cost function shifts the decision boundary during the gradient update phase to favor the minority class. By increasing the loss magnitude for misclassified stroke instances, the optimizer prioritizes the detection of positive cases over maximization of global accuracy. We performed stratified 5-fold patient-level cross-validation to evaluate model stability and generalization while preventing patient overlap between training and validation sets. Hyperparameters were selected within training folds using AUROC and AUPRC due to the imbalanced outcome distribution. The semantic and numerical branches were trained separately within each fold, and final performance was reported as mean $\pm$ standard deviation across folds. Data access requires approval and can be requested by contacting Authors~3 and~4.

\vspace{-2mm}
\begin{table*}[!t]
\centering
\caption{\small Baseline Characteristics of the Study Population Stratified by Stroke Status. P-values are calculated using Welch's t-test for continuous variables and Pearson's chi-squared test for categorical variables. SD: Standard Deviation; LDL: Low-Density Lipoprotein; HDL: High-Density Lipoprotein; A1c: Hemoglobin A1c; TG: Triglycerides; FBS: Fasting Blood Sugar.}
\label{tab:table1}
\resizebox{\textwidth}{!}{%
\begin{tabular}{lcccc}
\toprule
Characteristic & \makecell{Overall\\(N=102,736)} & \makecell{No Stroke\\(N=97,140)} & \makecell{Stroke\\(N=5,596)} & P-value \\
\midrule
Age (years), mean (SD) & 63.4 (13.8) & 62.9 (13.7) & 72.4 (11.1) & $<$0.001 \\
Systolic Blood Pressure (mmHg), mean (SD) & 129.2 (17.1) & 129.1 (17.1) & 130.9 (17.8) & $<$0.001 \\
Body Mass Index (kg/m$^2$), mean (SD) & 29.3 (6.6) & 29.4 (6.7) & 28.3 (5.3) & $<$0.001 \\
LDL Cholesterol (mmol/L), mean (SD) & 2.7 (1.0) & 2.8 (1.0) & 2.4 (1.0) & $<$0.001 \\
HDL Cholesterol (mmol/L), mean (SD) & 1.4 (0.4) & 1.4 (0.4) & 1.3 (0.4) & $<$0.001 \\
Hemoglobin A1c (\%), mean (SD) & 6.2 (0.9) & 6.2 (0.9) & 6.2 (0.8) & 0.736 \\
Triglycerides (mmol/L), mean (SD) & 1.5 (0.9) & 1.5 (0.9) & 1.5 (0.8) & 0.010 \\
Fasting Blood Sugar (mmol/L), mean (SD) & 5.8 (1.4) & 5.8 (1.4) & 5.9 (1.3) & $<$0.001 \\
Sex, n (\%) &  &  &  & $<$0.001 \\
Female & 63,848 (62.0) & 60,783 (95.2) & 3,065 (4.8) &  \\
Male & 38,888 (38.0) & 36,357 (93.5) & 2,531 (6.5) &  \\
Hypertension, n (\%) &  &  &  & $<$0.001 \\
No & 40,021 (39.0) & 38,744 (39.9) & 1,277 (22.8) &  \\
Yes & 62,715 (61.0) & 58,396 (60.1) & 4,319 (77.2) &  \\
Diabetes Mellitus, n (\%) &  &  &  & $<$0.001 \\
No & 71,439 (69.5) & 68,030 (70.0) & 3,409 (60.9) &  \\
Yes & 31,297 (30.5) & 29,110 (30.0) & 2,187 (39.1) &  \\
\bottomrule
\end{tabular}
}
\vspace{-4mm}
\end{table*}

\section{Results}
\label{results}

\begin{table*}[ht]
\centering
\caption{Performance comparison across different architectures and modalities.}
\label{tab:results}
\resizebox{0.7\textwidth}{!}{%
\begin{tabular}{l l c c c r c}
\toprule
\textbf{Architecture} & \textbf{Modality} & \textbf{AUROC} & \textbf{AUPRC} & \textbf{Brier} & \textbf{$\Delta$ AUC} & \textbf{p-value} \\
\midrule
Logistic Regression & Numerical    & $0.740 \pm 0.010$ & $0.098 \pm 0.010$ & 0.207 & $+0.0\%$  & --- \\
HGBM (Clinical)     & Numerical    & $0.720 \pm 0.011$ & $0.088 \pm 0.011$ & 0.156 & $-2.7\%$  & $0.0001$ \\
Bio-Transformer     & Textual      & $0.842 \pm 0.015$ & $0.344 \pm 0.044$ & 0.061 & $+13.7\%$ & $0.0001$ \\
MedGate Fusion      & Multi-modal  & $0.848 \pm 0.016$ & $0.346 \pm 0.054$ & 0.065 & $+14.5\%$ & $0.0003$ \\
\bottomrule
\end{tabular}}
\end{table*}

\begin{figure*}[ht]
    \centering
    % --- FIRST ROW: 3 FIGURES ---
    \begin{subfigure}[b]{0.31\textwidth}
        \centering
        \includegraphics[width=\textwidth]{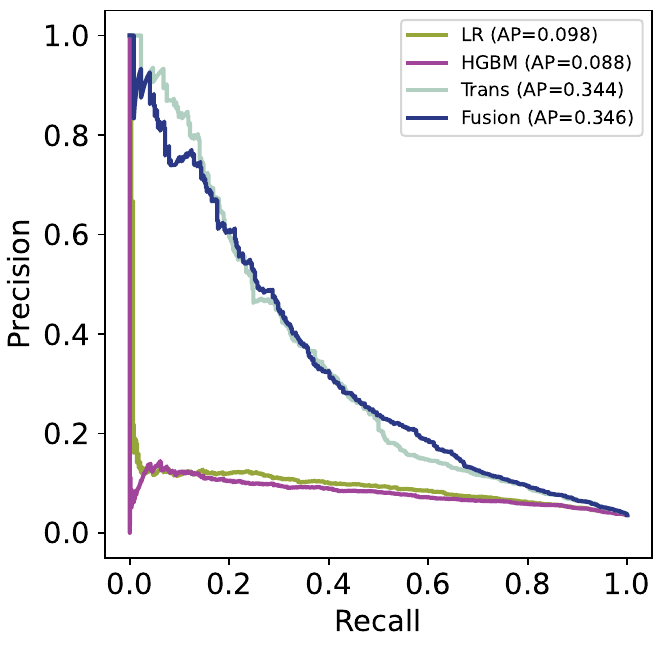}
        \caption{Precision-Recall}
        \label{fig:precision_recall}
    \end{subfigure}
    \hspace{.05cm}
    \begin{subfigure}[b]{0.335\textwidth}
        \centering
        \includegraphics[width=\textwidth]{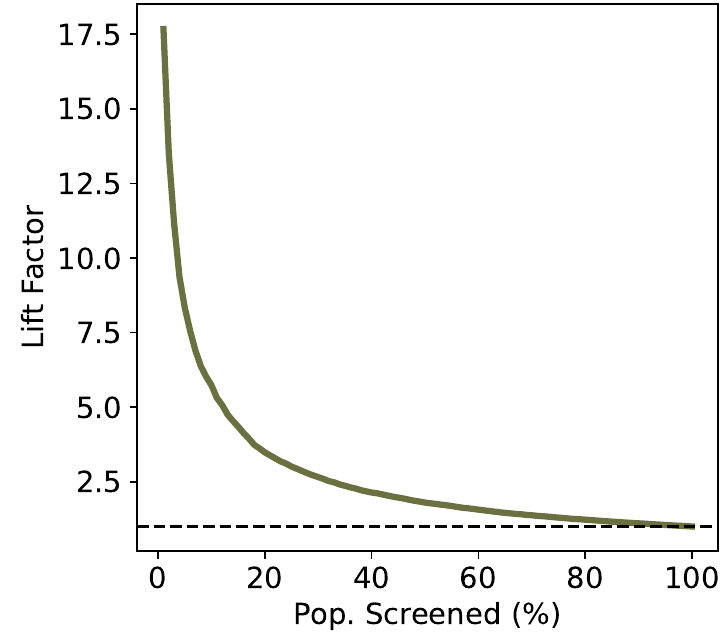}
        \caption{Clinical Lift}
        \label{fig:clinical_lift}
    \end{subfigure}
   \hspace{.05cm}
    \begin{subfigure}[b]{0.315\textwidth}
        \centering
        \includegraphics[width=\textwidth]{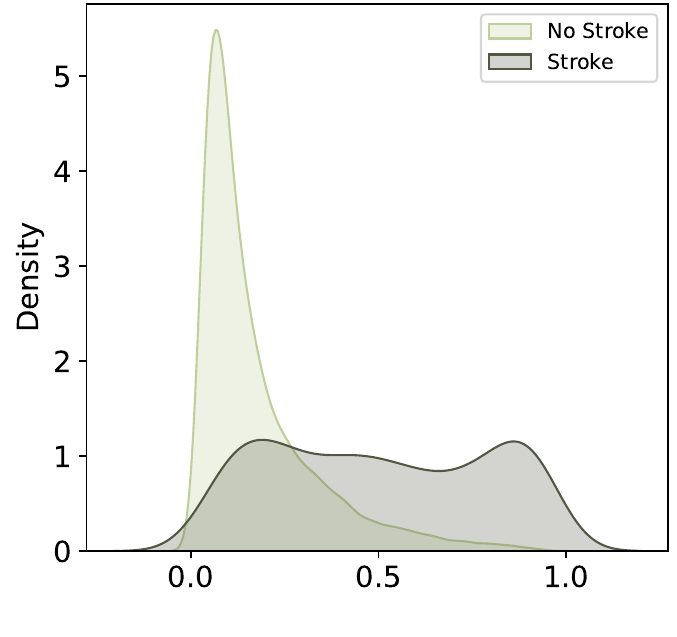}
        \caption{Separability}
        \label{fig:separability}
    \end{subfigure}

    \vspace{0.1cm} % Space between rows

    % --- SECOND ROW: 2 FIGURES ---
    \begin{subfigure}[b]{0.35\textwidth}
        \centering
        \includegraphics[width=\textwidth]{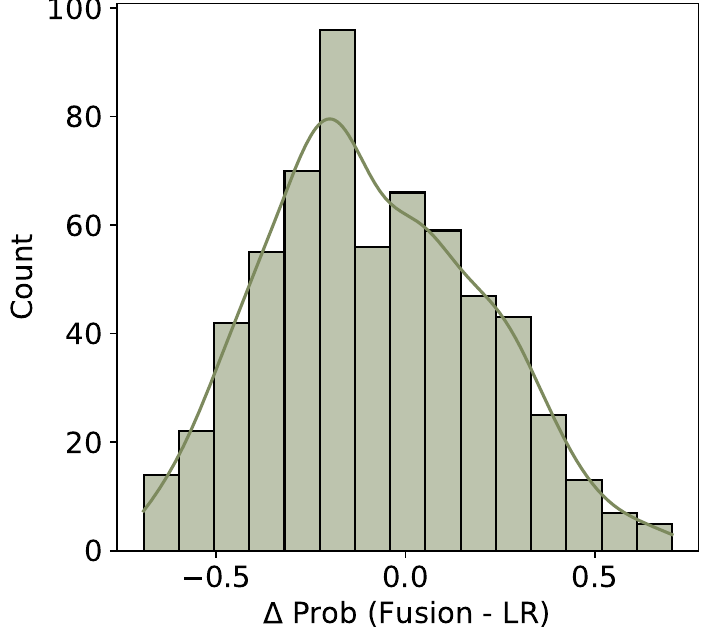}
        \caption{Marginal Gain}
        \label{fig:marginal_gain}
    \end{subfigure}
    \hspace{.1cm} % Adjust horizontal gap to center the bottom two
    \begin{subfigure}[b]{0.33\textwidth}
        \centering
        \includegraphics[width=\textwidth]{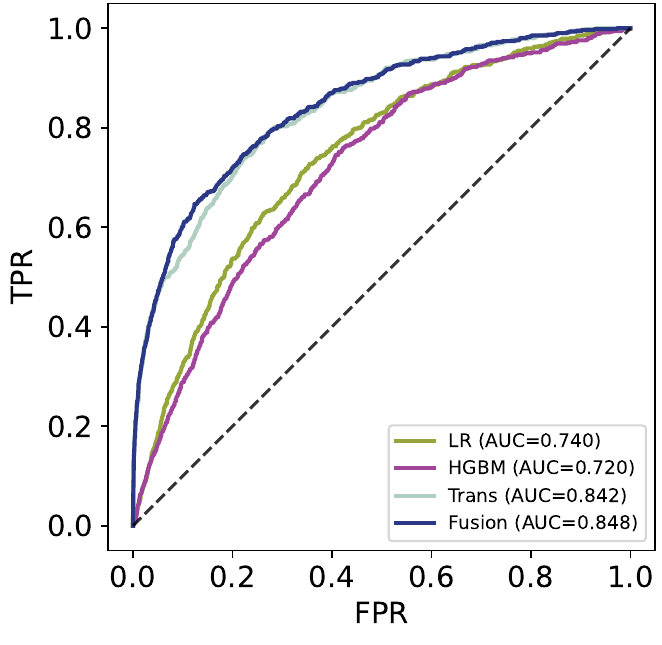}
        \caption{AUROC Comparison}
        \label{fig:auroc}
    \end{subfigure}

    \caption{\textbf{Benchmark Analysis.} Model performance comparison across clinical and textual modalities.}
    \label{fig:all}
    \vspace{-5mm}
\end{figure*}

\subsection{Predictive Performance and Statistical Benchmarking}

A prospective first-encounter protocol evaluated the discriminative power of the MedGate-Fusion architecture in five stratified cross-validation folds. The comparative results in Table~\ref{tab:results} show that the integration of multi-modal features provides a significant and stable performance advantage over traditional clinical baselines. The MedGate-Fusion model achieved an AUROC of 0.848 and an AUPRC of 0.346, providing the highest overall performance among the evaluated models. This performance represents a 14.5\% increase in AUROC relative to the baseline Logistic Regression. The statistical analysis confirms that the predictive gain is highly relevant and statistically significant ($p < 0.001$).
To evaluate the discriminative utility across the entire decision manifold, we generated the multi-model AUROC comparison in Figure~\ref{fig:auroc}. The ROC curves illustrate that the semantic transformer branch serves as the primary driver of predictive performance. The Bio-Transformer alone reached an AUROC of 0.842, suggesting that unstructured diagnostic narratives possess a high information density for future stroke risk. The numerical HGBM baseline attained a lower AUROC of 0.720, whereas fusing both modalities yields the most technical and robust risk manifold.

\subsection{Precision-Recall Dynamics and Decision Space Separability}

We analyzed the efficiency of the model in rare-event recovery through the Precision-Recall benchmarking presented in Figure~\ref{fig:precision_recall}. Standard numerical models often struggle with high class imbalance where the stroke prevalence is low. To address this challenge, the MedGate-Fusion model maintains a high and stable precision even at high levels of recall. The AUPRC of 0.346 suggests that the semantic signals in the first-encounter narratives are highly specific to vascular deterioration. As a result, the architecture provides a reliable tool for identifying high-risk individuals without incurring excessive false positives.

The separability of the decision space in Figure~\ref{fig:separability} visualizes the posterior probability density for the stroke and control cohorts. The MedGate-Fusion model effectively separates the two populations by pushing the stroke cases toward high-probability clusters. To quantify the specific contribution of the semantic branch, we examined the marginal gain in Figure~\ref{fig:marginal_gain}. The distribution of the probability delta ($\Delta$) illustrates that the addition of transformer-derived features significantly increases the risk estimation for true positive cases. The semantic narratives capture latent and complex risk patterns that traditional biomarkers ignore.

\subsection{Clinical Utility and Decile-based Lift Analysis}

A decile-based clinical lift analysis quantified the practical value of the proposed method, as shown in Figure~\ref{fig:clinical_lift}. This metric calculates the ratio between the model detection rate and the random screening baseline. The MedGate-Fusion model demonstrates a high and substantial lift factor in the top deciles of the screened population. For example, screening the highest 10\% of the identified risk group captures five times more stroke cases than random screening. The model therefore offers a high and relevant benefit for primary care intervention strategies.

We further evaluated the relationship between precision and sensitivity to establish the optimal clinical operating point. The calibration results in Table~\ref{tab:results} confirm that the predicted probabilities align with the observed event frequencies. The Brier score of 0.065 indicates a high and stable level of predictive reliability. As a result, the MedGate-Fusion architecture provides a reliable and scientifically valid framework for point-of-care decision support. Such metrics establish the feasibility of using the diagnostic text from the first-encounter as a primary evidence base for stroke prevention.

\section{Discussion}
\label{discussion}

\subsection{Semantic Information and Predictive Superiority}

To evaluate the predictive dominance of semantic features, we analyzed the performance of the Bio-Transformer branch relative to numerical baselines. The results in Table~\ref{tab:results} show that unstructured narratives provide a consistent signal for the risk of five-year stroke. This textual modality reached an AUROC of 0.842, exceeding the 0.720--0.740 range achieved by the numerical baselines. We hypothesize that diagnostic text captures longitudinal comorbidity trajectories that discrete physiological measurements cannot represent in a static laboratory snapshot \cite{li2022hi, amirahmadi2023deep}. This finding is consistent with recent literature suggesting that clinical notes contain latent pathological descriptors that act as early warning signals \cite{seinen2022use,rajkomar2018scalable,lyu2023multimodal}. These results suggest that first-encounter narratives encode clinically meaningful context (e.g., comorbidity burden, symptoms descriptions, and care patterns) that is not fully captured by a small set of structured markers in a single visit \cite{mahbub2022unstructured}. The strong performance of the text-only branch and the additional gain from fusion indicate complementary value rather than redundancy between modalities.

The discriminative power of the semantic modality was examined by computing the Information Gain $IG$ relative to the baseline distribution. The gain is defined as $IG(Y; X) = H(Y) - H(Y \mid X)$, where $H$ denotes the Shannon entropy of the stroke outcome. A high and positive $IG$ value for text indicates that clinical narratives reduce predictive uncertainty at the first visit.
The transformer architecture identifies high-risk sub-phenotypes that standard laboratory markers and physical measurements do not capture. These semantic cues provide the resolution needed to distinguish stable chronic patients from those near a vascular event. The textual branch thus functions as a contextual sensor of systemic pathological deterioration.

\subsection{Multi-modal Fusion and Decision Reliability}

The proposed MedGate-Fusion architecture integrates semantic and physiological modalities to verify the technical contribution of multimodal fusion. The improvement from the Bio-Transformer branch to MedGate-Fusion was modest in absolute AUROC terms, increasing from 0.842 to 0.848. This suggests that semantic narratives were the dominant modality in this cohort, while structured biomarkers added incremental discrimination. The value of fusion is therefore not only higher AUROC, but also a clinically grounded mechanism for anchoring narrative-derived risk to routinely measured physiological markers. We formulate the final fused risk score $S_f$ as a weighted combination $S_f = \alpha \cdot S_s + (1-\alpha) \cdot S_n$. In this equation, $S_s$ and $S_n$ denote semantic and numerical scores, while $\alpha$ represents the learned gating coefficient. To provide a stable and calibrated risk estimation, the model utilizes the gating mechanism to prioritize the most reliable modality. The fused manifold in Figure~\ref{fig:separability} shows a clear separation between the stroke and control cohorts.

To determine the impact of clinical biomarkers, we examined the marginal gain presented in Figure~\ref{fig:marginal_gain} for positive cases. The results indicate that laboratory values such as Hemoglobin A1c and systolic blood pressure provide a physical anchor for transformer embeddings. While the semantic signal is the primary driver, the numerical data prevents the model from relying solely on linguistic patterns. This finding confirms that the MedGate-Fusion architecture provides the most technical and robust risk manifold for primary care deployment. As a result, the model maintains a high and consistent level of performance even when diagnostic narratives are sparse or brief, while the fusion process balances the flexibility of deep learning with the stability of established biomarkers. Dictionary-based masking may also introduce a weak structural signal because the presence of masked tokens can reflect documentation patterns. Thus, we interpret the strong text-only performance cautiously and view fusion as evidence of substantial prognostic information in first-encounter narratives, rather than proof that fusion always yields large gains over text alone.

\subsection{Clinical Implications and Prospective Validity}

Figure~\ref{fig:clinical_lift} and Figure~\ref{fig:precision_recall} illustrate the decile-based lift used to quantify the clinical utility of the MedGate-Fusion model. As illustrated in these figures, the model reached an AUPRC of 0.346, which indicates a superior precision in a rare-event clinical environment \cite{saito2015precision,pinker2018reporting}. Our lift analysis suggests that screening the top 10\% of high-risk patients captures five times more cases than random screening. This level of detection outperforms the CHA2DS2-VASc score \cite{olesen2012value}, which typically yields AUROC values between 0.67 and 0.72 in recent studies. This architecture offers a practical tool for early intervention during the first clinical visit, where preventive management is most effective. The first-encounter protocol confirms that these predictions are prospective and valid for real-world primary care deployment in Canada.

The mitigation of leakages and the transition from identification to prediction were achieved employing a First-Encounter protocol. This methodological constraint confirms that the model identifies the early shadow of the stroke rather than the documented mirror of the event. The achieved AUROC of 0.848 shows that primary care narratives contain latent indicators that precede the formal diagnosis of a vascular event. In practice, the MedGate-Fusion model can support point-of-care risk stratification and earlier intervention. We also show that unstructured text alone can be sufficient for advanced predictive modeling in EMRs. Such findings provide a clear and technical path for integrating deep learning into standard clinical workflows for stroke prevention \cite{carlisle2026development}.

\subsection{EMR Limitations and Future Directions}

To provide a balanced evaluation, we identified limitations in the electronic medical record data and the current multi-center study design. Systematic missingness and professional coding variability often introduce noise into numerical biomarkers and unstructured diagnostic narratives in the CPSSN database. We recognized that the current Canadian cohort might exhibit regional documentation patterns that do not generalize to various healthcare systems. To address this, future research must validate the MedGate-Fusion architecture on global and multi-ethnic datasets to verify the performance. The current study is limited to a five-year prediction window, which may not capture the total lifetime risk for younger patients. We plan to extend the observation period to verify the longitudinal stability of the semantic and numerical risk signals.

Our work demonstrates that semantic screening offers a valid path for early stroke prevention strategies. However, the processing demands of semantic encoding may limit near-term adoption in small primary care settings. To address this, we plan to explore distilled and lighter transformer architectures to maintain performance while reducing the necessary hardware requirements. This research direction will verify if smaller semantic models can preserve the high information density found in the current study. This will support our goal of delivering a scalable risk screening solution that integrates seamlessly into existing healthcare IT. Such efforts will confirm that the benefits of multi-modal AI reach a broad and diverse patient population in Canada.

%%%%%%%%%%%%%%%%%%%%%%%%%%%%%%%%%%%%%%%%%%%%%%%%%%%%%%%%%%%%%%%%%%%%%%%%%%%%%%%%

\section{Conclusion}
\label{conclusion}

The MedGate-Fusion architecture proposed in this paper addresses the technical challenge of prospective stroke risk prediction in primary care settings. The evaluation processed 808{,}921 observations from the Canadian Primary Care Sentinel Surveillance Network repository. A reduction to 102{,}736 high-quality encounters yielded reliable information for model training. A true prospective window required the first clinical encounter for each unique patient. Results in Table~\ref{tab:results} confirm that unstructured diagnostic narratives contain predictive signals for vascular events.

The Bio-Transformer branch achieved an AUROC of 0.842 without lab measurements. A gated fusion mechanism integrated semantic embeddings and numerical biomarkers into a unified decision manifold. The MedGate-Fusion model reached an AUROC of 0.848 and an AUPRC of 0.346 on the test dataset. The marginal utility of numerical features suggests that diagnostic text captures core latent risk.

Semantic screening supports earlier stroke prevention in primary care. Clinical lift analysis indicates that the method captures five times more cases than random or standard screening protocols. Data missingness and regional coding variability remain primary limitations of this electronic medical record study. Future work should test generalizability across international and diverse clinical populations. The MedGate-Fusion architecture provides an automated solution for point-of-care risk stratification in Canada.

%%%%%%%%%%%%%%%%%%%%%%%%%%%%%%%%%%%%%%%%%%%%%%%%%%%%%%%%%%%%%%%%%%%%%%%%%%%%%%%%
% \section*{ACKNOWLEDGMENT}

% The authors would like to thank Dr. Karim Keshavjee for his assistance in facilitating access to the dataset used in this study.

%%%%%%%%%%%%%%%%%%%%%%%%%%%%%%%%%%%%%%%%%%%%%%%%%%%%%%%%%%%%%%%%%%%%%%%%%%%%%%%%

\printbibliography

\end{document}